%% file: cvpr.tex
\documentclass[10pt,twocolumn,letterpaper]{article}

\usepackage{cvpr}      
\input{preamble}
\definecolor{cvprblue}{rgb}{0.21,0.49,0.74}
\usepackage[pagebackref,breaklinks,colorlinks,allcolors=cvprblue]{hyperref}

\makeatletter
\def\@fnsymbol#1{\ifcase#1\or
  1\or
  2\or
  3\or
  4\or
  5\else
  \@arabic{#1}\fi}
\makeatother
\input{math_commands.tex}

\newcommand{\Rho}{\mathcal{P}}
\usepackage{multirow}

\begin{document}

\title{HiResNets: Native Full-HD Video Recognition with Foveal Residual Streams}

\author{
Shivani Mall \qquad
Swarnim Jain\footnotemark[1] \qquad
Jo\~{a}o F. Henriques\\
\begin{tabular}{c}
Visual Geometry Group, University of Oxford\\
{\tt\small shivanim@robots.ox.ac.uk}
\end{tabular}
}

\maketitle
\footnotetext[1]{Intern at VGG from the University of Cambridge.}

\input{sec/0_abstract}
\input{sec/1_intro}
\input{sec/2_formatting}

\input{sec/3_finalcopy_fixed}

\input{sec/fixed}
{
    \small
    \bibliographystyle{ieeenat_fullname}
    \bibliography{main}
}

\end{document}

%% file: math_commands.tex
\usepackage{amsmath,amsfonts,bm}

\def\eqref#1{equation~\ref{#1}}

\def\1{\bm{1}}

\DeclareMathAlphabet{\mathsfit}{\encodingdefault}{\sfdefault}{m}{sl}
\SetMathAlphabet{\mathsfit}{bold}{\encodingdefault}{\sfdefault}{bx}{n}



%% file: sec/0_abstract.tex
\begin{abstract}
Much of the recent progress in image and video recognition has come at the cost of memory: larger models, increased resolution, and longer temporal contexts.
An inevitable component is the quadratic (or larger) growth of memory and compute based on image resolution, which is a property of the grid sampling used in convolutional networks and vision transformers.
In this work we study residual networks whose convolutional blocks have logarithmic-square growth instead, enabling them to process very high-resolution video quickly.
The key insight is to use a residual architecture's residual stream as a high-resolution buffer, to which convolutional blocks only read and write via log-polar image warp operations.
Layers adaptively focus on different parts of each frame, with very high resolution only near the focus point.
A complete high-resolution representation is built up in the residual stream (theoretical construction presented to eliminate the quadratic dependency of stream resolution) which is analogous to eye saccades creating a complete picture in biological vision. Experiments demonstrate that our proposed HiResNets learn to foveate around scenes similarly to human vision, and have superior performance in difficult egocentric video recognition tasks, especially egocentric video with small objects and fine-grained recognition.
\end{abstract}

%% file: sec/1_intro.tex
\section{Introduction}
\label{intro}
Progress in image and video recognition has been driven by ever-larger models, higher input resolutions, and longer temporal contexts. This trend has clear costs: memory and compute grow at least quadratically with spatial resolution in convolutional networks and vision transformers \cite{dosovitskiy2021vit,he2016resnet}, creating a hard bottleneck for tasks that require fine detail. Applications such as egocentric video recognition or small-object analysis are especially constrained, not because of a lack of model capacity, but because of the inefficiency of uniform grid-based sampling.

\begin{figure*}[t]
    \centering
\includegraphics[width=\textwidth]{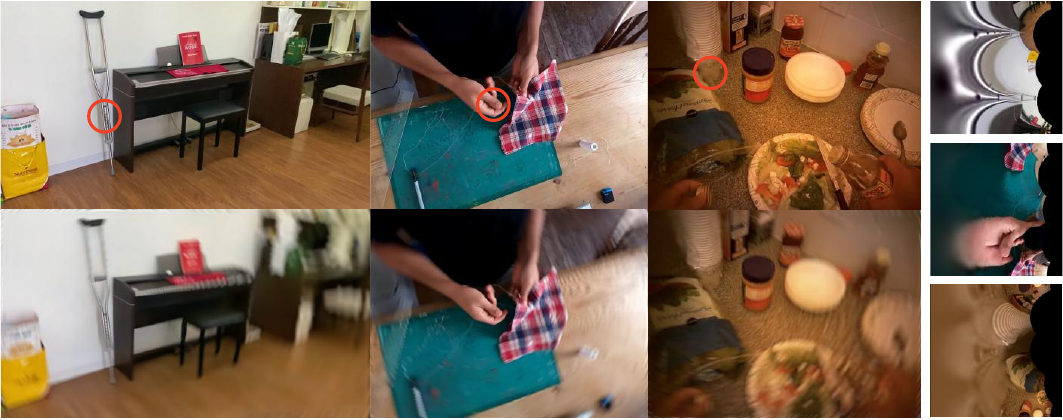}
    \\\hfill (a) EgoObjects \hspace{0.5cm} \hfill (b) PACO \hspace{4pt} \hfill (c) EGTEA \hspace{0.5cm} (d) Warps \hspace{1pt}\strut
    \caption{{\bf (a-c)} Illustration of non-uniform resolution on 3 datasets: EgoObjects, PACO and EGTEA. The first row shows original images and focus point (red circles), while the second row shows non-uniform resolution images (by warping and un-warping with a log-polar grid). Notice that regions away from the focus point (in the second row) have much less detail. {\bf (d)} The same 3 images, warped to log-polar space. The vertical axis corresponds to angular coordinates, and the horizontal axis to radial coordinates (distance from focus point). Toward the right of the image (far from the focus point), objects are compressed into smaller areas, and so a log-polar network will process them at much lower resolutions. Black pixels on the right are too far outside the warp grid.}
    \label{fig:splash}
\end{figure*}

In contrast, the human eye allocates resolution unevenly, capturing detail only at the fovea while encoding the periphery coarsely, and relies on saccades to integrate a full high-resolution scene \cite{meng2018foveated}. This principle has motivated several works: from earlier glimpse networks \cite{mnih2014ram}, zoom-in detectors \cite{wang2017zoomin}, and hierarchical multi-scale processing \cite{larochelle2010learning}, to more recent approaches which incorporated saccade-like glimpses into modern architectures, for example through recurrent hard-attention models \cite{elsayed2019saccader,pan2025mram} or differentiable foveated sampling schemes \cite{killick2023foveation}. Despite these advances, the common limitation is that foveation is applied \emph{outside the backbone itself} (typically as a control or pre-processing module wrapped around an otherwise standard network), so the backbone continues to scale quadratically with resolution.

We address this limitation by embedding foveation directly into the backbone. Our key idea is to use the residual pathway of a deep network as a \emph{persistent high-resolution buffer}, while convolutional blocks interact only with a warped, adaptive-resolution view. This view is produced by a log-polar image warp, which preserves fine detail near a chosen focus point and compresses the periphery \cite{schwartz1980retinotopy} (illustrated in fig. \ref{fig:splash}-(d)). As a result, the cost of residual blocks grows only logarithmically with resolution, rather than quadratically, while the residual stream maintains full fidelity. Layers can shift their focus adaptively across frames, gradually building up a complete high-resolution representation in a manner analogous to biological saccades.

While this means that the residual stream still scales quadratically with resolution, it is only involved in inexpensive copy and addition operations, and its back-propagation memory can be reduced (for example with gradient checkpointing \cite{chen2016trainingdeepnetssublinear}). We thus have greater performance gains by improving the residual blocks, which are often the bottleneck due to their expensive convolutions and expanded numbers of channels \cite{liu2022convnet2020s}.

The resulting architecture, \textbf{HiResNets}, integrates foveation directly into residual networks. Unlike prior approaches that wrap a standard backbone with glimpse or zoom modules, HiResNets \emph{modify the internal computation} of each residual block, yielding fundamentally different scaling behaviour with respect to input resolution. Our contributions are threefold:  
\begin{enumerate}
    \item A residual design in which block cost grows only logarithmically with resolution while the residual stream maintains full fidelity, along with a theoretical construction that eliminates the quadratic dependency of the stream's resolution.
    \item A differentiable log-polar warp mechanism enabling adaptive foveated processing inside the backbone itself.  
    \item Extensive experiments showing that HiResNets not only reduce memory and computation but also learn interpretable foveation strategies. 
\end{enumerate}
On egocentric video benchmarks, HiResNets offer consistent gains, particularly for fine-grained activities and small-object recognition.  
These results demonstrate that foveated architectures can overcome resolution bottlenecks in video understanding.

\begin{figure*}[t]
    \centering    \includegraphics[width=0.95\textwidth]{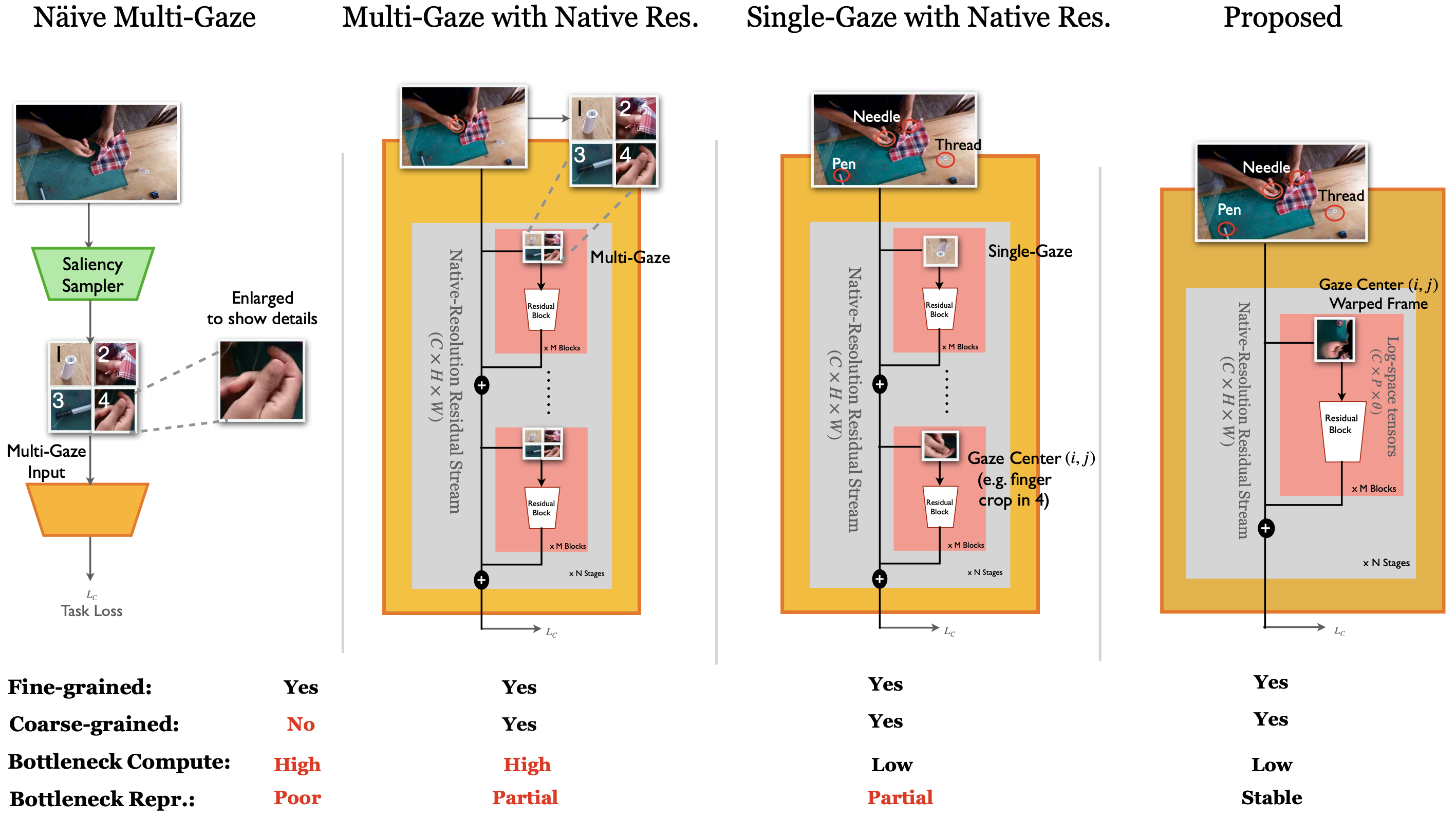}
    \caption{Overview of the differences between our scheme and alternative multi-gaze strategies. Using multi-gaze crops in every bottleneck block (column 2) leads to high compute, since all the gazed sites are preserved at high resolution. Using varying single-gaze crops instead (column 3), reduces compute but leads to degraded representations. Our proposal (column 4) preserves these single-gaze sites at high res., plus other frame regions at low res. This achieves an ideal balance between bottleneck compute and representations. Details in Sec. \ref{meth_diff}}
    \label{fig:diff}
\end{figure*}

%% file: sec/2_formatting.tex
\section{Related work} \label{sec:related}


There is an ample body of literature that is related to our work. In this section we provide only a small summary of more recent and classical papers that are relevant.

\paragraph{Biological Vision and Foveation.}
It is well known that human vision is sharply non-uniform: the retina contains a densely packed fovea surrounded by coarse peripheral sampling, with an organization that approximates a log-polar map. Classic studies of visual attention note that this arrangement is computationally efficient, as attention improves visual acuity only where it is needed \cite{carrasco2011attention}. Large variations in photoreceptor density across the retina are confirmed by anatomical evidence \cite{curcio1990photoreceptor}, and early computational models described this distribution as a log-polar coordinate transform \cite{schwartz1977retinotopy}. Perceptual research has shown how peripheral vision is limited by crowding and coarse feature integration \cite{strasburger2011peripheral}, while summary-statistic representations explain how robust search and whole-scene perception is possible despite these limits \cite{rosenholtz2012summary}.
Meng~et~al.~\cite{meng2018foveated} demonstrated computer vision applications of foveation, showing how uneven resolution sampling can reduce bandwidth and computation.
These works show that foveation and saccades evolved in biology as an efficient strategy for building complete high-resolution representations from partial glimpses, as opposed to having a uniform-resolution sensor, which is commonly the case in artificial vision.

\paragraph{Log-Polar Compression and Differentiable Warping.}
This biological perspective has inspired computer vision models that explicitly encode non-uniform sampling. Focusing on the deep learning era, early work in geometric warping introduced learnable modules such as spatial transformer networks (STN) \cite{jaderberg2015stn}, while polar transformer networks showed how polar coordinates yield built-in rotation and scale equivariance \cite{esteves2018polar}. More recently, log-polar convolution layers were proposed to natively operate in a retina-inspired coordinate system, yielding both efficiency and robustness to geometric variation \cite{su2022logpolar}.

\paragraph{Computational Models of Foveation.}
Beyond static reparameterizations, many learning-based systems attempt to mimic saccades by dynamically selecting where to process at high resolution. The Recurrent Model of Visual Attention (RAM) introduced sequential glimpses trained via reinforcement learning \cite{mnih2014ram}, while later models such as Saccader stabilized accuracy by supervising fixation selection \cite{elsayed2019saccader}. Other approaches replaced reinforcement learning with differentiable mechanisms: the Dynamic Zoom-In network, for example, predicted where to zoom within large images to save computation \cite{gao2018dynamiczoomin}.
Wang~et~al.~\cite{wang2017zoomin} introduced zoom-in detection pipelines, reducing cost for large images.
More recent methods have implemented continuous foveated sensors and learn how to shift fixations end-to-end \cite{killick2023foveation}, or incorporated foveation directly into transformers, as in FoveaTer \cite{jonnalagadda2021foveater}. Monte Carlo convolutions generalize filtering to non-uniform foveated inputs \cite{killick2022mcc}, and multi-resolution strategies such as Recently, Pan~et~al.~\cite{pan2025mram} proposed MRAM, a multi-level recurrent attention model that mimics fixations and saccades to improve stability and accuracy in glimpse-based architectures.

\paragraph{Egocentric Vision and Gaze Estimation Benchmarks.}
Egocentric video is a natural application domain for foveated models, since hand-object interactions, rapid egomotion, and small tools make uniform downsampling especially lossy. Benchmarks such as EPIC-KITCHENS \cite{damen2018epic} and Ego4D \cite{grauman2022ego4d} established large-scale testbeds for activity recognition and object understanding, while HD-EPIC \cite{perrett2025hdepic} recently added highly detailed annotations and gaze data.
In parallel, gaze-estimation datasets such as MPIIGaze \cite{zhang2019mpiigaze}, ETH-XGaze \cite{zhang2020ethxgaze}, and Gaze360 \cite{kellnhofer2019gaze360}, along with VR-focused corpora like OpenEDS \cite{garbin2019openeds}, provide evidence of where humans naturally focus in first-person settings. 

\paragraph{Small-Object Detection and Fine-Grained Targets.}
Standard detection backbones lose fine detail under pooling and stride, which led to several multi-scale architectures.
Larochelle and Hinton~\cite{larochelle2010learning} presented one of the earliest hierarchical multi-scale models, showing that learning across resolutions improves recognition efficiency.
Feature Pyramid Networks \cite{lin2017fpn} explicitly added top-down pathways to preserve detail across scales, while RetinaNet \cite{lin2017retinanet} introduced focal loss to mitigate the imbalance between small and large objects. 

\paragraph{Large-Scale High-Resolution Vision.}
Finally, a broad literature addresses how to scale vision architectures to high-resolution images and video. HRNet demonstrated the benefits of maintaining parallel high-resolution streams throughout a network \cite{sun2019hrnetpose,wang2020hrnetpami}, while Multiscale Vision Transformers (MViT) and its successor MViTv2 used hierarchical pooling to manage compute \cite{fan2021mvit,li2022mvitv2}. Efficiency-focused transformers prune or merge tokens dynamically, as in DynamicViT \cite{rao2021dynamicvit}, EViT \cite{liang2022evit}, and ToMe \cite{bolya2023tome}. At the extreme end, gigapixel pathology has motivated hierarchical pretraining (HIPT) \cite{chen2022hipt} and MIL-based slide classification (CLAM) \cite{lu2021clam}. However, these proposals mostly operate at globally fixed scales or discard fine detail. In contrast, the proposed HiResNets achieve sub-quadratic (log-squared) scaling by treating the residual stream itself as a high-resolution buffer accessed via log-polar warps, allowing full-HD video to be processed natively while preserving the biological analogy of foveation and saccades.

\begin{figure}[t]
    \centering    \includegraphics[width=\linewidth]{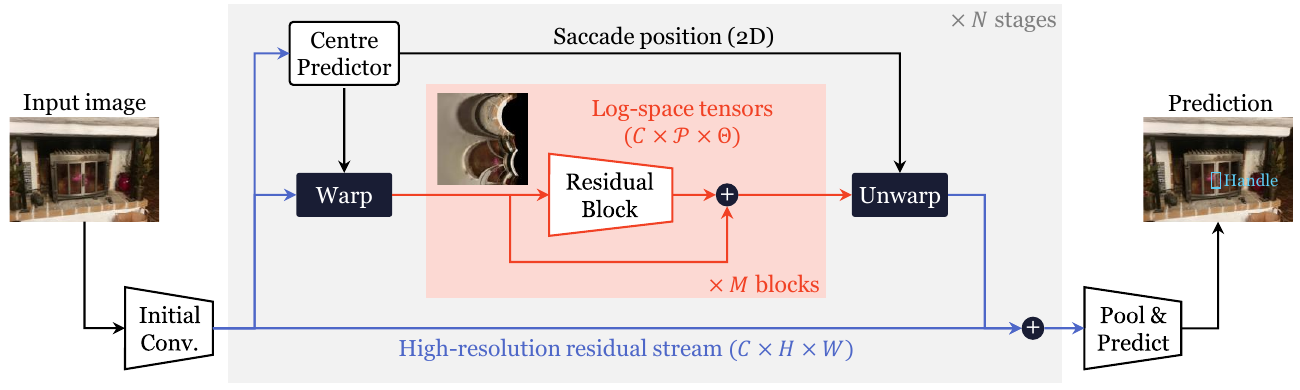}
    \caption{Overview of the proposed architecture. We integrate log-polar warp and unwarp operators around a series of residual (convolutional) blocks of a given backbone (e.g. ResNet). In this warped space, the network can process a very high resolution only around a center position (predicted by a separate branch), with much reduced computation. This allows the residual stream to carry information at a much higher resolution, processing full-HD video natively and recognizing very fine-grained detail.}
    \label{fig:pipeline}
\end{figure}

%% file: sec/3_finalcopy_fixed.tex
\section{Method}\label{sec:method}

\subsection{Residual networks and resolution scaling}
We begin by recalling the standard residual block notation of He~et~al.~\cite{he2016resnet}.  
Let $x^{(l)} \in \mathbb{R}^{C \times H \times W}$ denote the feature map (the residual stream) after block $l$.  
A residual block updates it via
\begin{equation}
    x^{(l+1)} = x^{(l)} + f\!\left(x^{(l)}\right),
\end{equation}
where $f(\cdot)$ is a sequence of convolutions, nonlinearities, and normalizations.  

While this formulation underlies most modern vision models, its cost grows quickly with resolution.  
The computational complexity of a convolution with kernel size $K$ is
\begin{equation}
    \mathcal{O}(C^2 H W K^2),
\end{equation}
so both compute and memory scale quadratically in the spatial dimensions.  
This quadratic law is the main bottleneck preventing standard residual networks from handling very high-resolution video.

\subsection{Log-polar warps}\label{sec:warp}
To alleviate this bottleneck, we require a representation that emphasizes local detail while compressing distant regions.
One candidate is the log-polar reparameterization of the image plane \cite{esteves2018polar}.  
Let $c \in \mathbb{R}^2$ be a centre of focus.  
The log-polar mapping $\psi_c : \mathbb{R}^{\Rho \times \Theta} \to \mathbb{R}^{H \times W}$ is defined by
\begin{equation}
    (i,j) = \psi_c(\rho,\theta) = c + \big(\exp(\rho)\cos\theta,\; \exp(\rho)\sin\theta\big),
\end{equation}
where $(\rho,\theta)$ are the log-polar coordinates of a point in the warped view.  
This mapping allocates exponentially more resolution near $c$, while progressively downsampling farther away.  

The warped feature at $(\rho,\theta)$ is obtained by bilinear resampling:
\begin{equation}
\begin{aligned}
v(\rho,\theta)
&=
\sum_{m}\sum_{n}
x_{m,n}\,
\varphi(i-m)\,
\varphi(j-n),\\
\varphi(t)
&=
\max(0,\,1-|t|).
\end{aligned}
\label{eq:bilinear}
\end{equation}
Since $\varphi$ has compact support in $[-1,1]$, this double sum reduces to exactly four terms: the 4 neighbours obtained by integer floor and ceiling of the coordinates.

\paragraph{Limitations of direct warping.}
Although the log-polar warp provides a more efficient parameterization, applying it directly to the input image or to the activations of a standard network has a major limitation: the centre $c$ is fixed for the entire forward pass.  
As a result, only a single part of the scene benefits from high-resolution processing, and the network cannot reallocate its resolution budget across layers or time.  
This motivates applying the warp at a more fine-grained level.

\subsection{HiResNet architecture}\label{sec:arch}
We therefore restructure the residual block so that convolutional processing takes place only in log-polar space, while the global high-resolution representation is preserved in the residual stream.  
Concretely, a block is defined as
\begin{equation}
    u = \psi_c(x), \qquad
    y = f(u), \qquad
    x \leftarrow x + \psi_c^{-1}(y),
\end{equation}
where $\psi_c^{-1}$ is the inverse log-polar warp, mapping features back to Cartesian coordinates by bilinear resampling.  

This means that each convolutional block operates on a compact warped view $u \in \mathbb{R}^{C \times \Rho \times \Theta}$, while $x$ maintains a full-resolution buffer of the scene.  
Over the course of $N$ blocks, the network integrates multiple warped updates into $x$, analogous to how the visual system fuses multiple saccades into a coherent high-resolution percept.
In practice, we let $f$ be a sequence of $M$ residual blocks instead of just one, in order to avoid recomputing the centre $c$ too often over the depth of the network.

\subsection{Center predictor}\label{sec:center-predictor}
To make this mechanism adaptive, the centre $c$ must be predicted dynamically at each block.  
We introduce a lightweight \emph{center predictor}, consisting of two $1\times 1$ convolutions with a ReLU nonlinearity, followed by a differentiable \emph{softargmax} operator \cite{warpedconv}.  
Given an attention map $a \in \mathbb{R}^{H \times W}$, the softargmax produces
\begin{equation}
    c = \sum_{i,j} (i,j)\, \frac{\exp(a_{i,j})}{\sum_{m,n}\exp(a_{m,n})}.
\end{equation}
This allows each block to reposition its high-resolution focus based on the current residual state $x$.

\subsection{Inverse warp}\label{sec:inverse}
In order to relate high-resolution information from different focus positions, we must be able to write information back from a log-polar warped space into a common space (typically cartesian).
We thus use also the following inverse warp to transform features to the high-resolution residual stream:
\begin{equation}
\begin{aligned}
(\rho,\theta)
&=
\psi_{c}^{-1}(i,j) \\
&=
\left(
\log\!\left\Vert(i',j')\right\Vert^{2},
\;
\mathrm{atan2}(j',i')
\right),\\
(i',j')
&=
(i,j)-c.
\end{aligned}
\end{equation}again using bilinear interpolation (eq. \ref{eq:bilinear}).

\subsection{Complexity analysis: from linear to logarithmic}\label{sec:analysis}
A standard residual block with kernel size $k$ applied to a feature map of size $(W, H)$ has cost $\mathcal{O}(k^2 W H)$. In our proposal, convolutions act only on the warped view $u \in \mathbb{R}^{C \times \Theta \times \Rho}$, where the log-polar warp yields $\Theta = \mathcal{O}(\log W)$ and $\Rho = \mathcal{O}(\log H)$. The warp itself can be implemented in $\mathcal{O}(\Theta \Rho)$, so the dominant cost is the convolution, scaling as
\[
\mathcal{O}(k^2 \Theta \Rho) = \mathcal{O}\!\big(k^2 \log W \, \log H \big).
\]
This logarithmic-square scaling replaces the quadratic growth of conventional blocks, and becomes the bottleneck whenever convolution dominates warp overhead and channel dimensions remain moderate.

Meanwhile, the residual stream ensures that no fine detail is lost: every update is reintegrated into a global high-resolution buffer.  
Thus HiResNets achieve logarithmic-square scaling in the computationally dominant convolutional bottlenecks, while preserving complete spatial information across depth and time.

%% file: sec/fixed.tex
\section{Implementation Details}

\label{sec:details}
In each experiment, we take an existing residual network as a baseline (ResNet \cite{he2016resnet} or SqueezeTime \cite{zhai2024timewastesqueezetime}), and obtain a HiResNet by adding the warp operations as described in sec. \ref{sec:arch}.
We implement the log-polar warp (sec. \ref{sec:warp}) and its inverse transform modules using efficient bilinear sampling (\texttt{grid\_sample} in PyTorch) in order to produce the warped and unwarped tensors.

A new localization network is instantiated in every stage (group of residual blocks) -- for example, the ResNet always has 4 stages \cite{he2016resnet}. Therefore there is one focus point per stage. We also experiment with the frequency of the polar prediction, for block groups with large number of residual blocks.

As for spatial resolution of the tensors, the residual blocks' spatial sizes increase by 1/2 every stage with increasing depth, while the residual stream has a constant size 4 times smaller than the input resolution.


\begin{figure}[t]
    \centering
    \includegraphics[width=0.95\linewidth]{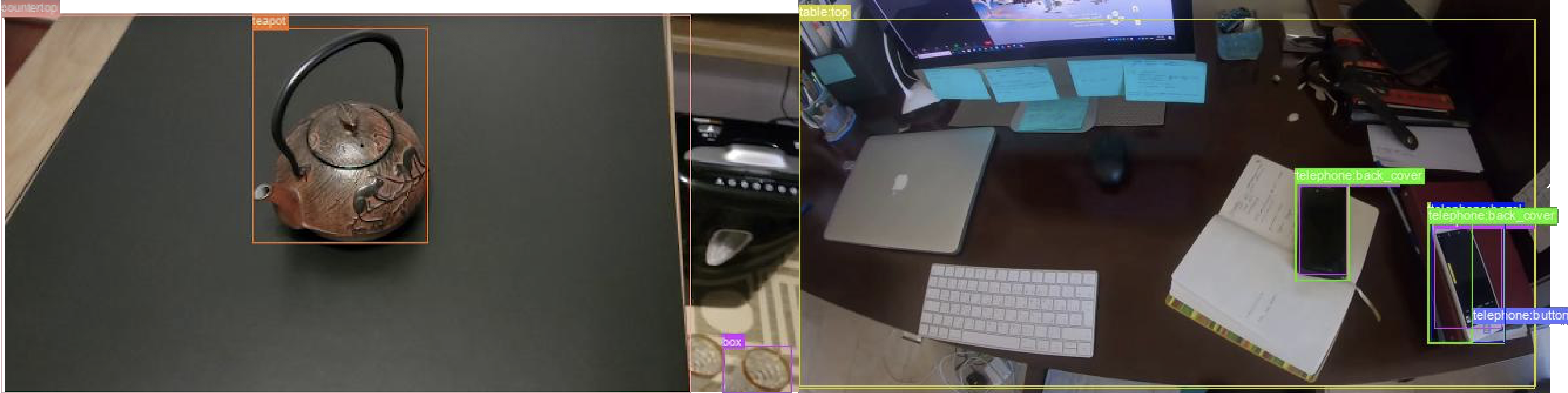}
    \\\hspace{0.02\textwidth} (a) \hfill\hfill (b) \hfill\strut
    \caption{Qualitative results (randomly selected): {\bf (a)} object detection on EgoObjects, {\bf (b)} object part detection on PACO. Notice that PACO includes fine-grained and small object details, such as the mobile phone button on the bottom-right of the image.}
    \label{fig:qualitative2}
\end{figure}

\section{Experiments} \label{sec:experiments}
In these experiments, we want to assess several capabilities of HiResNets: 1) their ability to mimic gaze estimation, analogously to biological vision; 2) the ability to process higher-resolution images and video than their corresponding baselines; 3) their performance scaling w.r.t. image resolution; 4) the ability to detect fine-grained object details that would be difficult in lower resolutions.

\begin{figure}[t]
    \centering
    \begin{subfigure}[t]{0.485\linewidth}
        \centering
        \includegraphics[width=\linewidth]{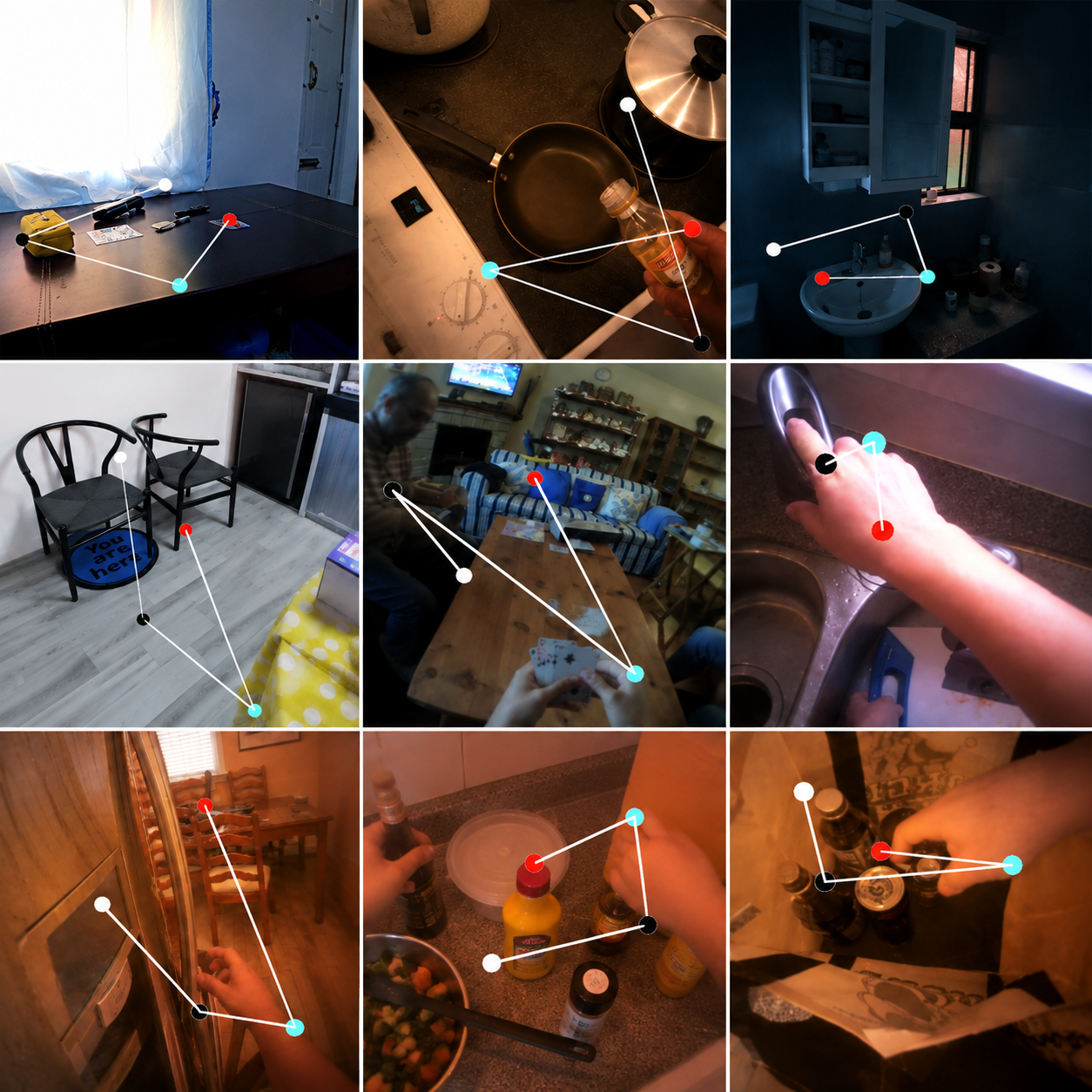}
        \caption{Qualitative results (randomly selected): gaze estimation on Ego4D and EGTEA. Notice that the network tends to focus on hands in close-up scenes (important for action prediction and locating manipulated objects), and alternating near-far regions in panoramic scenes.}
        \label{fig:qualitative1}
    \end{subfigure}
    \hfill
    \begin{subfigure}[t]{0.485\linewidth}
        \centering        \includegraphics[width=\linewidth]{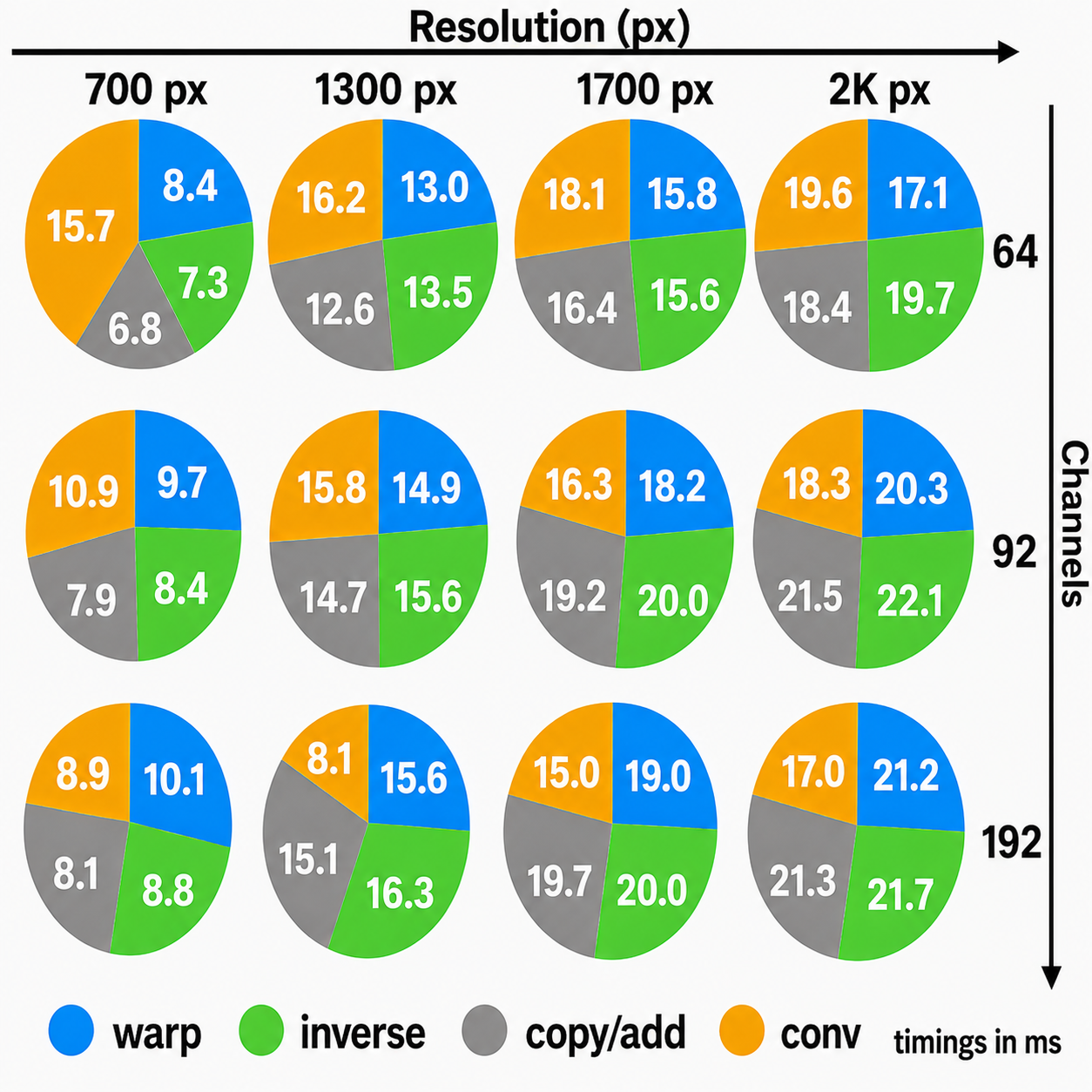}
        \caption{Per-block cost decomposition at four (700, 1.3K, 1.7K, 2K) resolutions and increasing channels (64, 92, 192). At higher resolutions, convolutions (conv.) take less of the total time, until they equal other operators (log-polar and inverse warp, residual stream copy/add). Log-square scaling applies to conv only.}
        \label{fig:timing}
    \end{subfigure}

    \caption{Overview of timing and qualitative gaze estimation results.}
    \label{fig:combined}
\end{figure}

We show examples of the tasks in figures \ref{fig:qualitative2} and \ref{fig:qualitative1}:
\begin{itemize}
    \item Gaze estimation in Ego4D and EGTEA;
    \item Object detection in EgoObjects and Ego4D;
    \item Object part detection in PACO.
\end{itemize}

\paragraph{Baselines.}
We use YOLOv11 \cite{yolo11_ultralytics} as a strong one-stage object detection baseline. It uses a CSPDarknet backbone with PANet feature aggregation and anchor-free detection heads, to trade off between accuracy and speed on high-resolution inputs.
For gaze estimation, Global-Local Correlation (GLC) \cite{lai2023eye} is a transformer-based egocentric gaze estimation model that explicitly models the interaction between global scene context and local visual features. It injects a ``global token'' into the transformer embedding and uses a Global-Local Correlation (GLC) module to compute attention weights between that global token and every local token.

\subsection{High-Resolution Egocentric Vision Experiments}

We focus on egocentric video, which presents unique challenges that are relevant for our approach. Unlike third-person data, egocentric data is dominated by rapid head motion, cluttered environments, and frequent occlusions. Objects of interest are often small, hand-held, and viewed at unusual angles. These conditions make accurate recognition heavily dependent on preserving fine spatial detail, but requiring high resolution quickly becomes prohibitive. We therefore focus our experiments on egocentric object detection, where the ability of HiResNets to foveate adaptively and process high-resolution video efficiently is most critical.

In all these experiments, we evaluate the method at different resolutions, including relatively high ones (reaching Full HD resolution), in order to show the scaling behaviour of the different methods.

\begin{table*}[t]
\centering

\begin{minipage}[t]{0.48\textwidth}
\centering
\scriptsize
\setlength{\tabcolsep}{3pt}
\renewcommand{\arraystretch}{0.9}

\begin{tabular}{lccccccc}
\toprule
 & & \multicolumn{3}{c}{EGTEA} & \multicolumn{3}{c}{Ego4D} \\
\cmidrule(lr){3-5}
\cmidrule(lr){6-8}
Method & Max Res. & AP & AR & F1 & AP & AR & F1 \\
\midrule

GLC \cite{lai2023eye}
& 640 & 35 & 61 & 44 & 34 & 57 & 43 \\
\cmidrule(lr){2-8}
& 1000 & 38 & 63 & 47 & 36 & 59 & 46 \\
\midrule

ARGaze \cite{argz}
& 640 & 40 & 66 & 50 & 39 & 62 & 48 \\
\cmidrule(lr){2-8}
& 1000 & 42 & 68 & 52 & 41 & 64 & 50 \\
\midrule

SqueezeTime \cite{zhai2024timewastesqueezetime}
& 640 & 29 & 59 & 39 & 32 & 57 & 41 \\
\cmidrule(lr){2-8}
& 1000 & 33 & 60 & 42 & 34 & 58 & 42 \\
\midrule

Ours
& 640 & \textbf{43} & \textbf{69} & \textbf{54}
& \textbf{42} & \textbf{65} & \textbf{52} \\
\cmidrule(lr){2-8}
& 1000 & \textbf{45} & \textbf{72} & \textbf{56}
& \textbf{44} & \textbf{67} & \textbf{54} \\
\bottomrule
\end{tabular}

\caption{Gaze estimation results on EGTEA and Ego4D.}
\label{table:Tab1}
\end{minipage}
\hfill
\begin{minipage}[t]{0.48\textwidth}
\centering
\scriptsize
\setlength{\tabcolsep}{3pt}
\renewcommand{\arraystretch}{0.9}

\begin{tabular}{lccccccc}
\toprule
 & & \multicolumn{3}{c}{Ego4D} & \multicolumn{3}{c}{Ego-Objects} \\
\cmidrule(lr){3-5}
\cmidrule(lr){6-8}
Method & Max Res. & Sm. & Md. & Lg. & Sm. & Md. & Lg. \\
\midrule

YOLOv11 \cite{yolo11_ultralytics}
& 640 & 40 & 20 & 25 & 33 & 20 & 32 \\
\cmidrule(lr){2-8}
& 900 & 42 & 22 & 27 & 35 & 22 & 34 \\
\cmidrule(lr){2-8}
& 1200 & 43 & 23 & 28 & 36 & 23 & 35 \\
\midrule

Ours
& 640 & \textbf{46} & \textbf{27} & \textbf{26}
& \textbf{39} & \textbf{25} & \textbf{33} \\
\cmidrule(lr){2-8}
& 900 & \textbf{50} & \textbf{29} & \textbf{28}
& \textbf{43} & \textbf{28} & \textbf{35} \\
\cmidrule(lr){2-8}
& 1200 & \textbf{51} & \textbf{30} & \textbf{29}
& \textbf{45} & \textbf{30} & \textbf{36} \\
\bottomrule
\end{tabular}

\caption{Object detection results on Ego4D and Ego-Objects. Accuracies reported on small (Sm.), medium (Md.), large (Lg.) splits.}
\label{table:Tab2}
\end{minipage}

\end{table*}

\paragraph{Results.}

For the the Ego4D dataset, when scaling the resolutions, accuracy does not significantly improve as seen in table \ref{table:Tab2}. This can be attributed to the distribution of the object sizes in the dataset which cover large pixel areas, hence do not need higher resolution training. In contrast, for EgoObjects, when scaling the resolutions, accuracy improves by about 10\%, attributing to varied object size distribution from small to large pixel areas (table \ref{table:Tab2}). However, we can observe that performance is generally higher at intermediate resolutions. Varying improvements at increasing resolutions across the two datasets motivates us to investigate performance on a dataset primarily covering smaller objects sizes. We show an example result for EgoObjects in fig. \ref{fig:qualitative2}-(a).

\subsection{Fine-grained Visual Understanding}

Finally, we turn to fine-grained visual understanding, for which we posit that high-resolution video understanding may be especially well-suited.
We thus evaluate on the PACO \cite{ramanathan2023paco} detection and classification dataset for detailed visual understanding. Derived from the Ego4D dataset \cite{grauman2022ego4d}, PACO comprises of rich annotations 531 object categories, of which 456 are object-part categories. There are a total of 15667 frames in train and 550 frames in validation, and 197K bounding boxes. 
Here, we also study the effectiveness of foveation by evaluating on dataset splits with varying object sizes (spatial area 0-5\% and 5-10\% of the image). While our method delivers stable evaluation performance for these splits, the baseline's accuracy degrades by 1-2\% than that observed in table \ref{table:Tab3}.


\paragraph{Results.}
We show an example result in fig. \ref{fig:qualitative2}-(b), which illustrates the difficulties posed by the small parts in this dataset (such as the mobile phone button in the bottom-right corner).
We evaluate the performance on PACO when scaling resolution from 640 to 1400. We see that our method outperforms the baseline at lower resolutions, however, only by a small margin (table \ref{table:Tab3}). This indicates that PACO is not solvable with resolution lower than its full-HD or higher. The performance increase at higher pixel resolutions (that is, 900 and 1400) comes at 1.2x and 1.5x higher latency respectively, compared to the 640 pixel resolution.
This shows that, for harder classification and detection tasks, increased resolution does meaningfully improve performance, and methods such as ours can take advantage of the additional detail without a large increase in computational cost. We observe 5 \% accuracy improvement over the ResNet \cite{he2016resnet} baseline in the image classification task, and showing example qualitative results in Fig \ref{fig:qualitative1}-(a). More detailed classification results are in the appendix.


\begin{figure}[t]
    \centering
    \begin{subfigure}[t]{0.49\textwidth}
        \centering        \includegraphics[width=0.55\linewidth]{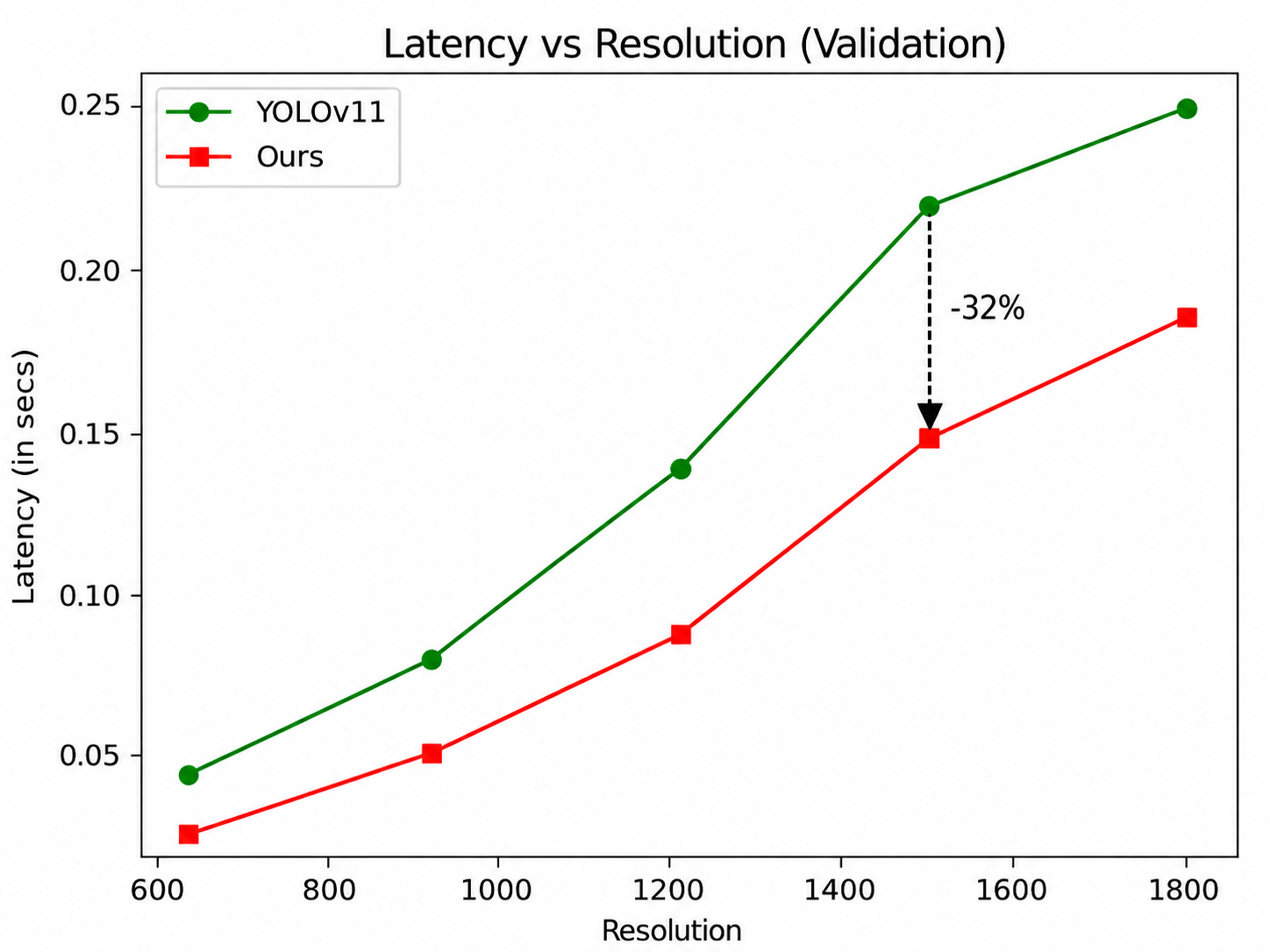}      \label{fig:cc_lat_val}
    \end{subfigure}
    \hfill
    \begin{subfigure}[t]{0.49\textwidth}
        \centering        \includegraphics[width=0.55\linewidth]{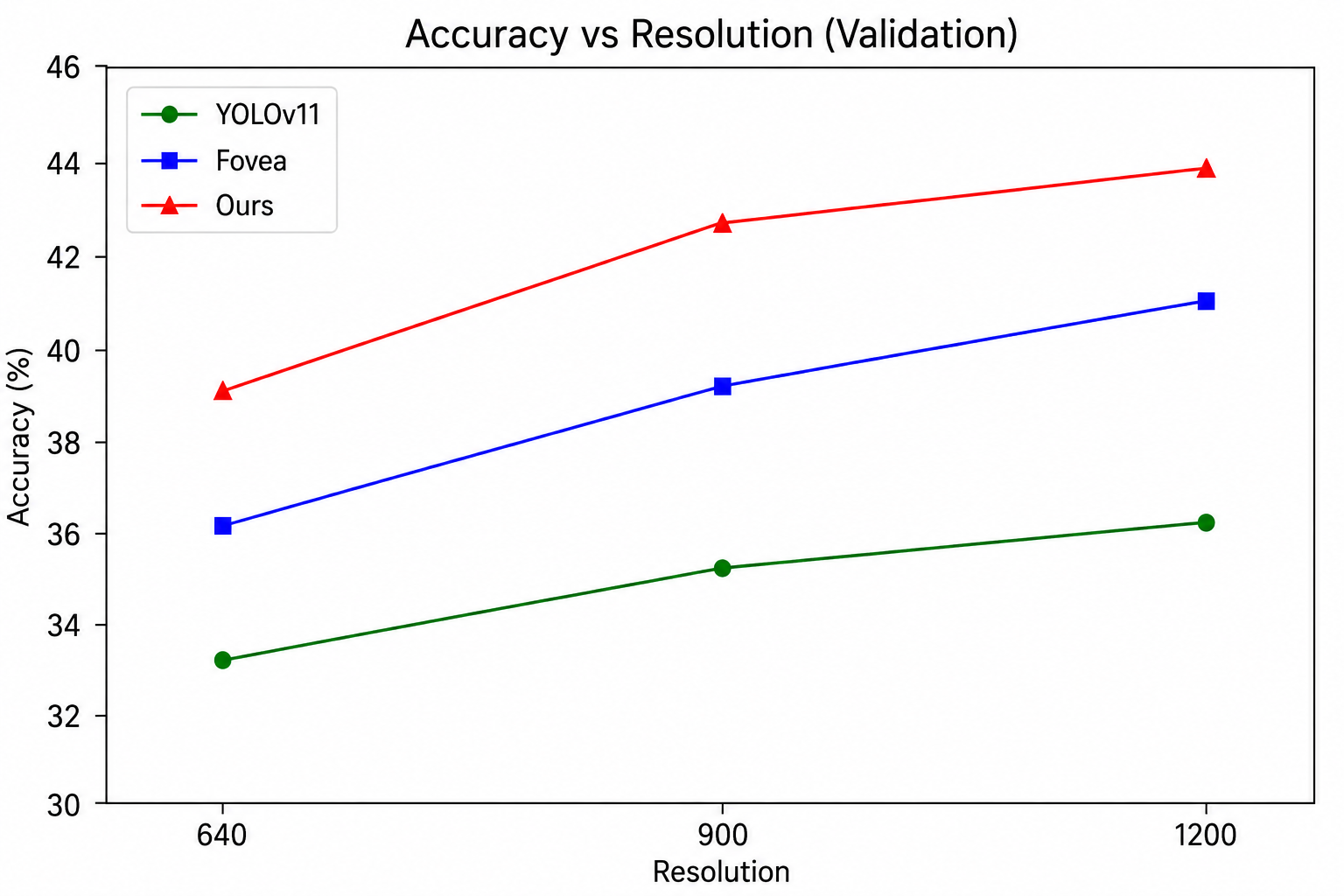}
        
        \label{fig:paco_acc_resolution}
    \end{subfigure}

    \caption{
(a) Latency vs.\ resolution on Ego4D dataset. Linearity dominates in our plot as the log-square scaling applies to conv. only, while other operators are still linear
(b) Accuracy vs.\ resolution on the small Ego-Objects with YOLOv11, FOVEA, and ours. 
}
    \label{fig:resolution_comparison}
\end{figure}

\subsection{Scaling behaviour of computational cost with increasing image resolution}

For this experiment we measured inference latency as a function of input resolution on the Ego4D gaze estimation benchmark, without any additional training. We compared HiResNets to the baseline SqueezeTime, running both models on an NVIDIA M40 GPU. Latency was recorded while scaling the input from a typical ResNet resolution (224px) up to 1K, using identical preprocessing and batch (size 24) settings.

\paragraph{Results.} The results, shown in Fig. \ref{fig:timing} and \ref{fig:resolution_comparison}, show that HiResNets scale much more favorably with resolution. SqueezeTime, FOVEA and YOLOv11 exhibits the expected quadratic growth in latency as image size increases, reflecting the pixel-wise scaling of convolutional layers in a ResNet-style backbone. In contrast, HiResNets grow only nearly linearly with resolution, consistent with their logarithmic–squared property.
As a result, HiResNets remain efficient even at 1K input, while SqueezeTime becomes substantially slower.



\begin{table*}[t]
\centering
\scriptsize
\setlength{\tabcolsep}{1pt}
\renewcommand{\arraystretch}{0.85}

\begin{minipage}[t]{0.54\textwidth}
\centering
\begin{tabular*}{\linewidth}{@{\extracolsep{\fill}}lcccccc@{}}
\toprule
Method & Res. & Train & Eval. & AP & AR \\
\midrule
YOLOv11 \cite{yolo11_ultralytics} 
& 640  & 50 & 42 & 56 & 18 \\
& 900  & 54 & 45 & 58 & 21 \\
& 1400 & 58 & 47 & 59 & 23 \\
\midrule
Ours
& 640  & \bf{56} & \bf{47} & \bf{62} & \bf{27} \\
& 900  & \bf{60} & \bf{50} & \bf{64} & \bf{30} \\
& 1400 & \bf{64} & \bf{53} & \bf{66} & \bf{33} \\
\bottomrule
\end{tabular*}
\caption{Object part detection results on PACO.}
\label{table:Tab3}
\end{minipage}
\hfill
\begin{minipage}[t]{0.40\textwidth}
\centering
\setlength{\tabcolsep}{1pt}

\begin{tabular}{lcc}
\toprule
Method & AP$\uparrow$ & Lat.$\downarrow$ \\
\midrule
Learning to Zoom \cite{salzoom} & 39.7 & 9.5 \\
Dynamic Conv. \cite{dyconv} & 41.1 & 10.4 \\
Deformable Conv. \cite{defconv} & 42.3 & 11.6 \\
\textbf{Ours} & \textbf{46.1} & \textbf{7.5} \\
\bottomrule
\end{tabular}

\caption{CNN variant comparison on Egtea.}
\label{table:tab4}
\end{minipage}

\end{table*}
\section{Discussion}

\subsection{Bottleneck Latency vs Representational Capacity Trade-off} \label{disc}
The center predictor (Sec. \ref{sec:center-predictor}) with the respective transform operation (inverse log-polar or log-polar) is an instantiation of the STN \cite{jaderberg2015stn}. As described in \cite{warpedconv, jaderberg2015stn}, this fast self-contained module can be inserted into a CNN architecture at any point, and in any number; plus, the CNN can be designed with various transforms, in parallel. We wrap our backbone's bottleneck blocks with this module. In fact, several other works have also explored the benefits from input-dependent geometric transformations of feature maps \cite{defconv, dyconv}. Here, we perform a comparative study with two such plug-in modules -- deformable and dynamic convolutions. In table \ref{table:tab4}, we show the varying impact on the bottleneck's feature processing latency and representational capacity.

\subsection{Gaze-Shift or Dynamic Saccades: An Emergent Property}
The dataset gaze trajectories naturally capture gaze-shift or dynamic saccades from the observer \cite{li2018egtea, grauman2022ego4d}. We observe a higher distance correlation between the ground-truth and predicted gaze trajectories in ours and baseline methods, except Learning to Zoom \cite{salzoom}. This shows that gaze shift emerges as a result of applying input-dependent spatial transformation modules (Sec. \ref{disc}) throughout the architecture backbone which is absent in \cite{salzoom}. See Fig. \ref{fig:qualitative2} for predicted gaze trajectories from our method.

\subsection{Residual Stream Memory Consumption} 
We also propose a theoretical construction that eliminates the quadratic dependency of the residual stream resolution (while introducing a quadratic dependency on the depth). While the residual stream is full resolution ($O(n^2)$), we prove that this buffer can be eliminated entirely. It is true that, as written, the residual stream $x \in \mathbb{R}^{C \times H \times W}$ scales quadratically with resolution. We will better discuss this limitation of a direct implementation. 

Since $x$ is only ever written to by unwarped log-polar updates, we have $x = \sum_{k<l} \psi_{c_k}^{-1}(y_k)$ at block $l$, so the block's input is $u_l = \psi_{c_l}(x) = \sum_{k<l} \psi_{c_l} \circ \psi_{c_k}^{-1}(y_k)$. Defining the direct log-polar-to-log-polar warp $W_{k\to l} \triangleq \psi_{c_l} \circ \psi_{c_k}^{-1}$ (a single warp with a closed-form grid), the block update becomes $u_l = \sum_{k<l} W_{k\to l}(y_k)$. This is mathematically equivalent to the original formulation but never instantiates $x$: only the log-polar tensors $\{y_k\} \in \mathbb{R}^{C \times \Rho \times \Theta}$ are stored.

In the current implementation, we also see that the memory growth is within reasonable budget as resolution increases (only upto 10 GB at 4K px)

\subsection{HiResNets Roadmap}
\label{meth_diff}
Gaze is an underlying property of egocentric videos, primarily obtained by following an agent's eye gaze. To preserve salient frame regions at high-resolution, gaze serves as a crucial cue. Thus, we propose to predict the gaze center within a frame, following a top-down visual mechanism (also present in human vision). To arrive at our proposal, we ablate with a few baselines. For the first two, we preserve high-resolution frame crops around possible gaze centers (predicted), and jointly process these at each bottleneck block (Fig. \ref{fig:diff}, columns a, b). However, this leads to a high compute cost at each block. Additionally, successive gaze centers do not iteratively refine upon the previous ones. So, instead in our last baseline (Fig. \ref{fig:diff}, column c), we predict a fresh gaze center at every stage (set of bottleneck blocks), using feedback signal from the previous predictions; we then process a high-resolution crop at the recent gaze center only, and observe significantly reduced compute cost.

\subsection{Accuracy Analysis: Fine vs Coarse Regions}

We use high resolution data to primarily benefit fine-grained regions (e.g., in small objects) where details are inherently lost. We still maintain low resolution data in the bottleneck blocks for coarse-grained regions (e.g., large objects), in fact, preventing overfitting. Observe the improved accuracy on small objects while no degradation on larger ones (Fig. \ref{fig:resolution_comparison} and Table \ref{table:Tab2}). Related works \cite{fovea, salisa} discuss this challenge or trade-off as well, and like in our proposal, these methods also do not see a performance boost in coarse-grained regions (e.g., objects covering large pixel areas).

\section{Ablations for HiResNet Design}

As discussed in Sec. \ref{disc}, the center predictor (with the respective warp) is an instantiation of the STN, and can be inserted in a CNN architecture at any point and in any number \cite{warpedconv, jaderberg2015stn}. This combined module augments static convolutions by enabling spatial transformations of the input feature map; thus, having multiple of these, through a network's increasing depths is not an adverse effect of our or previous proposals. In fact, as discussed in \cite{warpedconv, jaderberg2015stn}, such input-adaptive modules lead to a higher representational alignment in the architecture's progressive stages, that is otherwise not possible in general CNNs. To corroborate this finding, we experiment with different configurations that result from predicting fresh centers for the transform through the network's depth. For this module, we further justify how we arrive at the position of insertion into the CNN architecture.

\subsection{Number of Center Predictors}
We experiment with increasing the center predictions, by predicting twice (instead of once) for stacks of residual blocks (greater than 4), and observe an increase in evaluation accuracy by 1.5\%. This is expected due to the representations drift in later blocks from that of the initial residual block. We also ablate by predicting the center only once for the entire network, which results in accuracy degradation ($>7\%$). Further, we ablate with using ground-truth image center (no prediction) which leads to an even more severe performance hit.

\subsection{Position of Center Predictors} As discussed in Sec. \ref{disc}, this module can be inserted in a CNN architecture at any point. We ablate with inserting the module before, instead of within a bottleneck block (as proposed). We observe severely lower accuracy as the full-fidelity frames are no longer preserved in the residual stream, leading to degraded representations through the network depth.



\section{Conclusion} \label{sec:conclusion}
In this work we propose HiResNets, a residual architecture that integrates log-polar warps into the residual stream to enable efficient foveated processing. Our experiments show that HiResNets can mimic gaze allocation, handle inputs at substantially higher resolutions than conventional baselines, and exhibit favourable scaling behaviour with respect to resolution. These properties lead to superior performance on egocentric tasks, where capturing fine-grained details is critical. Overall, our results highlight foveated representations as a promising direction for building scalable and efficient high-resolution vision models without incurring quadratic computational cost.

%% file: cvpr.bbl
\begin{thebibliography}{53}
\providecommand{\natexlab}[1]{#1}
\providecommand{\url}[1]{\texttt{#1}}
\expandafter\ifx\csname urlstyle\endcsname\relax
  \providecommand{\doi}[1]{doi: #1}\else
  \providecommand{\doi}{doi: \begingroup \urlstyle{rm}\Url}\fi

\bibitem[Bejnordi et~al.(2022)Bejnordi, Habibian, Porikli, and Ghodrati]{salisa}
Babak~Ehteshami Bejnordi, Amirhossein Habibian, Fatih Porikli, and Amir Ghodrati.
\newblock Salisa: Saliency-based input sampling for efficient video object detection, 2022.

\bibitem[Bolya et~al.(2023)Bolya, Fu, Dai, Zhang, Feichtenhofer, and Hoffman]{bolya2023tome}
Daniel Bolya, Cheng-Yang Fu, Xiaoliang Dai, Peizhao Zhang, Christoph Feichtenhofer, and Judy Hoffman.
\newblock Token merging: Your vit but faster.
\newblock In \emph{International Conference on Learning Representations (ICLR)}, 2023.

\bibitem[Carrasco(2011)]{carrasco2011attention}
Marisa Carrasco.
\newblock Visual attention: The past 25 years.
\newblock \emph{Vision Research}, 51\penalty0 (13):\penalty0 1484--1525, 2011.

\bibitem[Chen et~al.(2022)Chen, Chen, Li, Chen, Trister, Krishnan, and Mahmood]{chen2022hipt}
Richard~J. Chen, Chengkuan Chen, Yicong Li, Tiffany~Y. Chen, Andrew~D. Trister, Rahul~G. Krishnan, and Faisal Mahmood.
\newblock Scaling vision transformers to gigapixel images via hierarchical self-supervised learning.
\newblock In \emph{Proceedings of the IEEE/CVF Conference on Computer Vision and Pattern Recognition (CVPR)}, 2022.

\bibitem[Chen et~al.(2016)Chen, Xu, Zhang, and Guestrin]{chen2016trainingdeepnetssublinear}
Tianqi Chen, Bing Xu, Chiyuan Zhang, and Carlos Guestrin.
\newblock Training deep nets with sublinear memory cost, 2016.

\bibitem[Chen et~al.(2020)Chen, Dai, Liu, Chen, Yuan, and Liu]{dyconv}
Yinpeng Chen, Xiyang Dai, Mengchen Liu, Dongdong Chen, Lu Yuan, and Zicheng Liu.
\newblock Dynamic convolution: Attention over convolution kernels.
\newblock In \emph{CVPR}, 2020.

\bibitem[Curcio et~al.(1990)Curcio, Sloan, Kalina, and Hendrickson]{curcio1990photoreceptor}
Christine~A. Curcio, Kenneth~R. Sloan, Richard~E. Kalina, and Alan~E. Hendrickson.
\newblock Human photoreceptor topography.
\newblock \emph{Journal of Comparative Neurology}, 292\penalty0 (4):\penalty0 497--523, 1990.

\bibitem[Dai et~al.(2017)Dai, Qi, Xiong, Li, Zhang, Hu, and Wei]{defconv}
Jifeng Dai, Haozhi Qi, Yuwen Xiong, Yi Li, Guodong Zhang, Han Hu, and Yichen Wei.
\newblock Deformable convolutional networks.
\newblock In \emph{ICCV}, 2017.

\bibitem[Damen et~al.(2018)Damen, Doughty, Farinella, Fidler, Furnari, Kazakos, Moltisanti, Munro, Perrett, Price, and Wray]{damen2018epic}
Dima Damen, Hazel Doughty, Giovanni~Maria Farinella, Sanja Fidler, Antonino Furnari, Evangelos Kazakos, Davide Moltisanti, Jonathan Munro, Toby Perrett, Will Price, and Michael Wray.
\newblock Scaling egocentric vision: The {EPIC-KITCHENS} dataset.
\newblock In \emph{European Conference on Computer Vision (ECCV)}, 2018.

\bibitem[Dosovitskiy et~al.(2021)Dosovitskiy, Beyer, Kolesnikov, Weissenborn, Zhai, Unterthiner, Dehghani, Minderer, Heigold, Gelly, Uszkoreit, and Houlsby]{dosovitskiy2021vit}
Alexey Dosovitskiy, Lucas Beyer, Alexander Kolesnikov, Dirk Weissenborn, Xiaohua Zhai, Thomas Unterthiner, Mostafa Dehghani, Matthias Minderer, Georg Heigold, Sylvain Gelly, Jakob Uszkoreit, and Neil Houlsby.
\newblock An image is worth 16x16 words: Transformers for image recognition at scale.
\newblock In \emph{International Conference on Learning Representations}, 2021.

\bibitem[Elsayed et~al.(2019)Elsayed, Kornblith, and Le]{elsayed2019saccader}
Gamaleldin~F Elsayed, Simon Kornblith, and Quoc~V Le.
\newblock {Saccader}: Improving accuracy of hard attention models for vision.
\newblock In \emph{Advances in Neural Information Processing Systems}, 2019.

\bibitem[Esteves et~al.(2018)Esteves, Allen-Blanchette, Zhou, and Daniilidis]{esteves2018polar}
Carlos Esteves, Christine Allen-Blanchette, Xiaowei Zhou, and Kostas Daniilidis.
\newblock Polar transformer networks.
\newblock In \emph{International Conference on Learning Representations}, 2018.

\bibitem[Fan et~al.(2021)Fan, Xiong, Mangalam, Li, Yan, Malik, and Feichtenhofer]{fan2021mvit}
Haoqi Fan, Bo Xiong, Karttikeya Mangalam, Yanghao Li, Zhicheng Yan, Jitendra Malik, and Christoph Feichtenhofer.
\newblock Multiscale vision transformers.
\newblock In \emph{Proceedings of the IEEE/CVF International Conference on Computer Vision (ICCV)}, 2021.

\bibitem[Gao et~al.(2018)Gao, Yu, Li, Morariu, and Davis]{gao2018dynamiczoomin}
Mingfei Gao, Ruichi Yu, Ang Li, Vlad~I. Morariu, and Larry~S. Davis.
\newblock Dynamic zoom-in network for fast object detection in large images.
\newblock In \emph{Proceedings of the IEEE/CVF Conference on Computer Vision and Pattern Recognition (CVPR)}, 2018.

\bibitem[Garbin et~al.(2019)Garbin, Shen, Goodman, Lagergren, and Talathi]{garbin2019openeds}
Stephan~J. Garbin, Yiru Shen, Dushyant Goodman, Jakob~H. Lagergren, and Sachin~S. Talathi.
\newblock {OpenEDS}: Open eye dataset.
\newblock \emph{arXiv preprint arXiv:1905.03702}, 2019.

\bibitem[Grauman et~al.(2022)]{grauman2022ego4d}
Kristen Grauman et~al.
\newblock {Ego4D}: Around the world in 3,000 hours of egocentric video.
\newblock In \emph{Proceedings of the IEEE/CVF Conference on Computer Vision and Pattern Recognition (CVPR)}, 2022.

\bibitem[He et~al.(2016)He, Zhang, Ren, and Sun]{he2016resnet}
Kaiming He, Xiangyu Zhang, Shaoqing Ren, and Jian Sun.
\newblock Deep residual learning for image recognition.
\newblock In \emph{Proceedings of the IEEE conference on computer vision and pattern recognition}, pages 770--778, 2016.

\bibitem[Henriques and Vedaldi(2017)]{warpedconv}
Joao~F Henriques and Andrea Vedaldi.
\newblock Warped convolutions: Efficient invariance to spatial transformations.
\newblock In \emph{International Conference on Machine Learning}, pages 1461--1469. PMLR, 2017.

\bibitem[Jaderberg et~al.(2015)Jaderberg, Simonyan, Zisserman, and Kavukcuoglu]{jaderberg2015stn}
Max Jaderberg, Karen Simonyan, Andrew Zisserman, and Koray Kavukcuoglu.
\newblock Spatial transformer networks.
\newblock In \emph{Advances in Neural Information Processing Systems}, 2015.

\bibitem[Jocher and Qiu(2024)]{yolo11_ultralytics}
Glenn Jocher and Jing Qiu.
\newblock Ultralytics yolo11, 2024.

\bibitem[Jonnalagadda et~al.(2021)Jonnalagadda, Wang, Manjunath, and Eckstein]{jonnalagadda2021foveater}
Aditya Jonnalagadda, William~Yang Wang, B.S. Manjunath, and Miguel~P. Eckstein.
\newblock {FoveaTer}: Foveated transformer for image classification.
\newblock \emph{arXiv preprint arXiv:2105.14173}, 2021.

\bibitem[Kellnhofer et~al.(2019)Kellnhofer, Recasens, Stent, Matusik, and Torralba]{kellnhofer2019gaze360}
Petr Kellnhofer, Adri{\`a} Recasens, Simon Stent, Wojciech Matusik, and Antonio Torralba.
\newblock {Gaze360}: Physically unconstrained gaze estimation in the wild.
\newblock In \emph{Proceedings of the IEEE/CVF International Conference on Computer Vision (ICCV)}, 2019.

\bibitem[Killick et~al.(2022)Killick, Aragon-Camarasa, and Siebert]{killick2022mcc}
George Killick, Gerardo Aragon-Camarasa, and J.~Paul Siebert.
\newblock Monte-carlo convolutions on foveated images.
\newblock In \emph{Proceedings of the 17th International Joint Conference on Computer Vision, Imaging and Computer Graphics Theory and Applications (VISIGRAPP) -- VISAPP}, 2022.

\bibitem[Killick et~al.(2023)Killick, Henderson, Siebert, and Aragon-Camarasa]{killick2023foveation}
George Killick, Paul Henderson, Jan~Paul Siebert, and Gerardo Aragon-Camarasa.
\newblock Foveation in the era of deep learning.
\newblock In \emph{British Machine Vision Conference (BMVC)}, 2023.

\bibitem[Lai et~al.(2023)Lai, Liu, Ryan, and Rehg]{lai2023eye}
Bolin Lai, Miao Liu, Fiona Ryan, and James~M Rehg.
\newblock In the eye of transformer: Global--local correlation for egocentric gaze estimation and beyond.
\newblock \emph{International Journal of Computer Vision}, pages 1--18, 2023.

\bibitem[Larochelle and Hinton(2010)]{larochelle2010learning}
Hugo Larochelle and Geoffrey~E Hinton.
\newblock Learning to combine foveal glimpses with a third-order boltzmann machine.
\newblock In \emph{Advances in Neural Information Processing Systems}, 2010.

\bibitem[Li et~al.(2026)Li, Zhao, Deng, Lai, Wu, Chen, Froehlich, Zhao, and Tian]{argz}
Jia Li, Wenjie Zhao, Shijian Deng, Bolin Lai, Yuheng Wu, Ruijia Chen, Jon~E. Froehlich, Yuhang Zhao, and Yapeng Tian.
\newblock Autoregressive transformers for online egocentric gaze estimation, 2026.

\bibitem[Li et~al.(2018)Li, Liu, and Rehg]{li2018egtea}
Yin Li, Miao Liu, and James~M. Rehg.
\newblock In the eye of the beholder: Joint learning of gaze and actions in first person video.
\newblock In \emph{European Conference on Computer Vision (ECCV)}, 2018.

\bibitem[Li et~al.(2022)Li, Wu, Fan, Mangalam, Xiong, Malik, and Feichtenhofer]{li2022mvitv2}
Yanghao Li, Chao-Yuan Wu, Haoqi Fan, Karttikeya Mangalam, Bo Xiong, Jitendra Malik, and Christoph Feichtenhofer.
\newblock {MViTv2}: Improved multiscale vision transformers for classification and detection.
\newblock In \emph{Proceedings of the IEEE/CVF Conference on Computer Vision and Pattern Recognition (CVPR)}, 2022.

\bibitem[Liang et~al.(2022)Liang, Ge, Tong, Song, Wang, and Xie]{liang2022evit}
Youwei Liang, Chongjian Ge, Zhan Tong, Yibing Song, Jue Wang, and Pengtao Xie.
\newblock Evit: Expediting vision transformers via token reorganizations.
\newblock In \emph{International Conference on Learning Representations (ICLR)}, 2022.

\bibitem[Lin et~al.(2017{\natexlab{a}})Lin, Doll{\'a}r, Girshick, He, Hariharan, and Belongie]{lin2017fpn}
Tsung-Yi Lin, Piotr Doll{\'a}r, Ross~B. Girshick, Kaiming He, Bharath Hariharan, and Serge Belongie.
\newblock Feature pyramid networks for object detection.
\newblock In \emph{Proceedings of the IEEE/CVF Conference on Computer Vision and Pattern Recognition (CVPR)}, 2017{\natexlab{a}}.

\bibitem[Lin et~al.(2017{\natexlab{b}})Lin, Goyal, Girshick, He, and Doll{\'a}r]{lin2017retinanet}
Tsung-Yi Lin, Priya Goyal, Ross~B. Girshick, Kaiming He, and Piotr Doll{\'a}r.
\newblock Focal loss for dense object detection.
\newblock In \emph{Proceedings of the IEEE/CVF International Conference on Computer Vision (ICCV)}, 2017{\natexlab{b}}.

\bibitem[Liu et~al.(2022)Liu, Mao, Wu, Feichtenhofer, Darrell, and Xie]{liu2022convnet2020s}
Zhuang Liu, Hanzi Mao, Chao-Yuan Wu, Christoph Feichtenhofer, Trevor Darrell, and Saining Xie.
\newblock A convnet for the 2020s.
\newblock In \emph{Proceedings of the IEEE conference on computer vision and pattern recognition}, 2022.

\bibitem[Lu et~al.(2021)Lu, Williamson, Chen, Chen, Barbieri, and Mahmood]{lu2021clam}
Ming~Y. Lu, Drew~F.K. Williamson, Tiffany~Y. Chen, Richard~J. Chen, Matteo Barbieri, and Faisal Mahmood.
\newblock Data-efficient and weakly supervised computational pathology on whole-slide images.
\newblock \emph{Nature Biomedical Engineering}, 5:\penalty0 555--570, 2021.

\bibitem[Meng et~al.(2018)Meng, Du, Zwicker, and Varshney]{meng2018foveated}
Xiaoxu Meng, Ruofei Du, Matthias Zwicker, and Amitabh Varshney.
\newblock Kernel foveated rendering.
\newblock \emph{Proceedings of the ACM on Computer Graphics and Interactive Techniques}, 1:\penalty0 1--20, 2018.

\bibitem[Mnih et~al.(2014)Mnih, Heess, Graves, and Kavukcuoglu]{mnih2014ram}
Volodymyr Mnih, Nicolas Heess, Alex Graves, and Koray Kavukcuoglu.
\newblock Recurrent models of visual attention.
\newblock In \emph{Advances in Neural Information Processing Systems}, 2014.

\bibitem[Pan et~al.(2025)Pan, Yonekura, and Kuniyoshi]{pan2025mram}
Pengcheng Pan, Shogo Yonekura, and Yasuo Kuniyoshi.
\newblock \emph{Emergence of Fixational and Saccadic Movements in a Multi-level Recurrent Attention Model for Vision}, page 299–313.
\newblock Springer Nature Singapore, 2025.

\bibitem[Perrett et~al.(2025)Perrett, Darkhalil, Sinha, Emara, Pollard, Parida, Liu, Gatti, Bansal, Flanagan, Chalk, Zhu, Guerrier, Abdelazim, Zhu, Moltisanti, Wray, Doughty, and Damen]{perrett2025hdepic}
Toby Perrett, Ahmad Darkhalil, Saptarshi Sinha, Omar Emara, Sam Pollard, Kranti Parida, Kaiting Liu, Prajwal Gatti, Siddhant Bansal, Kevin Flanagan, Jacob Chalk, Zhifan Zhu, Rhodri Guerrier, Fahd Abdelazim, Bin Zhu, Davide Moltisanti, Michael Wray, Hazel Doughty, and Dima Damen.
\newblock {HD-EPIC}: A highly-detailed egocentric video dataset.
\newblock In \emph{Proceedings of the IEEE/CVF Conference on Computer Vision and Pattern Recognition (CVPR)}, 2025.

\bibitem[Ramanathan et~al.(2023)Ramanathan, Kalia, Petrovic, Wen, Zheng, Guo, Wang, Marquez, Kovvuri, Kadian, Mousavi, Song, Dubey, and Mahajan]{ramanathan2023paco}
Vignesh Ramanathan, Anmol Kalia, Vladan Petrovic, Yi Wen, Baixue Zheng, Baishan Guo, Rui Wang, Aaron Marquez, Rama Kovvuri, Abhishek Kadian, Amir Mousavi, Yiwen Song, Abhimanyu Dubey, and Dhruv Mahajan.
\newblock Paco: Parts and attributes of common objects.
\newblock In \emph{Proceedings of the IEEE/CVF Conference on Computer Vision and Pattern Recognition (CVPR)}, 2023.

\bibitem[Rao et~al.(2021)Rao, Zhao, Liu, Lu, Zhou, and Hsieh]{rao2021dynamicvit}
Yongming Rao, Wenliang Zhao, Benlin Liu, Jiwen Lu, Jie Zhou, and Cho-Jui Hsieh.
\newblock Dynamicvit: Efficient vision transformers with dynamic token sparsification.
\newblock In \emph{Advances in Neural Information Processing Systems}, 2021.

\bibitem[Recasens et~al.(2018)Recasens, Kellnhofer, Stent, Matusik, and Torralba]{salzoom}
Adri\`a Recasens, Petr Kellnhofer, Simon Stent, Wojciech Matusik, and Antonio Torralba.
\newblock Learning to zoom: a saliency-based sampling layer for neural networks.
\newblock In \emph{European Conference on Computer Vision (ECCV)}, 2018.

\bibitem[Rosenholtz et~al.(2012)Rosenholtz, Huang, Raj, Balas, and Ilie]{rosenholtz2012summary}
Ruth Rosenholtz, Jie Huang, Alvin Raj, Benjamin~J. Balas, and Livia Ilie.
\newblock A summary statistic representation in peripheral vision explains visual search.
\newblock \emph{Journal of Vision}, 12\penalty0 (4):\penalty0 14--14, 2012.

\bibitem[Schwartz(1977)]{schwartz1977retinotopy}
Eric~L. Schwartz.
\newblock Spatial mapping in the primate sensory projection: Analytic structure and relevance to perception.
\newblock \emph{Biological Cybernetics}, 25\penalty0 (4):\penalty0 181--194, 1977.

\bibitem[Schwartz(1980)]{schwartz1980retinotopy}
Eric~L Schwartz.
\newblock Computational anatomy and functional architecture of striate cortex: A spatial mapping approach to perceptual coding.
\newblock \emph{Vision Research}, 20\penalty0 (8):\penalty0 645--669, 1980.

\bibitem[Strasburger et~al.(2011)Strasburger, Rentschler, and J{\"u}ttner]{strasburger2011peripheral}
Hans Strasburger, Ingo Rentschler, and Martin J{\"u}ttner.
\newblock Peripheral vision and pattern recognition: A review.
\newblock \emph{Journal of Vision}, 11\penalty0 (5):\penalty0 13--13, 2011.

\bibitem[Su and Wen(2022)]{su2022logpolar}
Bing Su and Ji-Rong Wen.
\newblock Log-polar space convolution layers.
\newblock In \emph{Advances in Neural Information Processing Systems}, 2022.

\bibitem[Sun et~al.(2019)Sun, Xiao, Liu, and Wang]{sun2019hrnetpose}
Ke Sun, Bin Xiao, Dong Liu, and Jingdong Wang.
\newblock Deep high-resolution representation learning for human pose estimation.
\newblock In \emph{Proceedings of the IEEE/CVF Conference on Computer Vision and Pattern Recognition (CVPR)}, 2019.

\bibitem[Thavamani et~al.(2021)Thavamani, Li, Cebron, and Ramanan]{fovea}
Chittesh Thavamani, Mengtian Li, Nicolas Cebron, and Deva Ramanan.
\newblock Fovea: Foveated image magnification for autonomous navigation.
\newblock In \emph{Proceedings of the IEEE/CVF International Conference on Computer Vision (ICCV)}, pages 15519--15528, 2021.

\bibitem[Wang et~al.(2020)Wang, Sun, Cheng, Jiang, Deng, Zhao, Liu, Mu, Tan, Wang, Liu, and Xiao]{wang2020hrnetpami}
Jingdong Wang, Ke Sun, Tianheng Cheng, Borui Jiang, Chaorui Deng, Yang Zhao, Dong Liu, Yadong Mu, Mingkui Tan, Xinggang Wang, Wenyu Liu, and Bin Xiao.
\newblock Deep high-resolution representation learning for visual recognition.
\newblock \emph{IEEE Transactions on Pattern Analysis and Machine Intelligence}, 2020.

\bibitem[Wang et~al.(2017)Wang, Yin, Shi, Fang, Li, and Wang]{wang2017zoomin}
Zhe Wang, Yanxin Yin, Jianping Shi, Wei Fang, Hongsheng Li, and Xiaogang Wang.
\newblock Zoom-in-net: Deep mining lesions for diabetic retinopathy detection.
\newblock In \emph{Medical Image Computing and Computer Assisted Intervention -- MICCAI 2017}, pages 267--275, 2017.

\bibitem[Zhai et~al.(2024)Zhai, Li, Tang, Chen, and Wang]{zhai2024timewastesqueezetime}
Yingjie Zhai, Wenshuo Li, Yehui Tang, Xinghao Chen, and Yunhe Wang.
\newblock No time to waste: Squeeze time into channel for mobile video understanding, 2024.

\bibitem[Zhang et~al.(2019)Zhang, Sugano, Fritz, and Bulling]{zhang2019mpiigaze}
Xucong Zhang, Yusuke Sugano, Mario Fritz, and Andreas Bulling.
\newblock {MPIIGaze}: Real-world dataset and deep appearance-based gaze estimation.
\newblock \emph{IEEE Transactions on Pattern Analysis and Machine Intelligence}, 41\penalty0 (1):\penalty0 162--175, 2019.

\bibitem[Zhang et~al.(2020)Zhang, Park, Beeler, Bradley, Tang, and Hilliges]{zhang2020ethxgaze}
Xucong Zhang, Seonwook Park, Thabo Beeler, Derek Bradley, Siyu Tang, and Otmar Hilliges.
\newblock {ETH-XGaze}: A large scale dataset for gaze estimation under extreme head pose and gaze variation.
\newblock In \emph{European Conference on Computer Vision (ECCV)}, 2020.

\end{thebibliography}
